\documentclass[10pt, a4paper]{article}
\usepackage{lrec2022} 
\usepackage{multibib}
\newcites{languageresource}{Language Resources}
\usepackage{graphicx}
\usepackage{tabularx}
\usepackage{soul}
\usepackage{titlesec}
\titleformat{\section}{\normalfont\large\bfseries\center}{\thesection.}{1em}{}
\titleformat{\subsection}{\normalfont\SmallTitleFont\bfseries\raggedright}{\thesubsection.}{1em}{}
\titleformat{\subsubsection}{\normalfont\normalsize\bfseries\raggedright}{\thesubsubsection.}{1em}{}
\renewcommand\thesection{\arabic{section}}
\renewcommand\thesubsection{\thesection.\arabic{subsection}}
\renewcommand\thesubsubsection{\thesubsection.\arabic{subsubsection}}

\usepackage{epstopdf}
\usepackage[utf8]{inputenc}

\usepackage{hyperref}
\usepackage{xstring}

\usepackage{color}

\usepackage{multirow}
\usepackage{graphicx}
\usepackage{float}

\usepackage{booktabs}

\title{Electoral Agitation Data Set: The Use Case of the Polish Election}

\name{Mateusz Baran, Mateusz Wójcik, Piotr Kolebski, \\
{\bf \large Michał Bernaczyk, Krzysztof Rajda, Łukasz Augustyniak, Tomasz Kajdanowicz 
     \vspace{0.25em} }} 

\address{
    Wrocław University of Science and Technology \\
    Department of Computational Intelligence, Wrocław, Poland \\
     \{mateusz.baran, mateusz.wojcik, piotr.kolebski\}@pwr.edu.pl\\
     \{krzysztof.rajda, lukasz.augustyniak, tomasz.kajdanowicz\}@pwr.edu.pl
     \vspace{0.25em}\\
    Wrocław University \\ 
    Faculty of Law, Administration and Economics, Wrocław, Poland \\
     michal.bernaczyk@uwr.edu.pl
}

\abstract{
The popularity of social media makes politicians use it for political advertisement. Therefore, social media is full of electoral agitation (electioneering), especially during the election campaigns. The election administration cannot track the spread and quantity of messages that count as agitation under the election code. It addresses a crucial problem, while also uncovering a niche that has not been effectively targeted so far. Hence, we present the first publicly open data set for detecting electoral agitation in the Polish language. It contains 6,112 human-annotated tweets tagged with four legally conditioned categories. We achieved a 0.66 inter-annotator agreement (Cohen’s kappa score). An additional annotator resolved the mismatches between the first two improving the consistency and complexity of the annotation process. The newly created data set was used to fine-tune a Polish Language Model called HerBERT (achieving a 68\% F1 score). We also present a number of potential use cases for such data sets and models, enriching the paper with an analysis of the Polish 2020 Presidential Election on Twitter.
 \\ \newline \Keywords{Electoral Agitation, Data Set, HerBERT, Natural Language Processing} }

\begin{document}

\maketitleabstract

\section{Introduction}

The use of social media in politics varies between two extremes: from creating an ethically debatable illusion of mass support to the brutal fight against political opponents in coordinated hate campaigns \cite{Kearney2013}. The scope of identified actions ranges from mass trolling, hate-speech, harassment, and intimidation to the spread of manipulative, defamatory, or false content (fake news) \cite{Skogerb2015,rashkin2017truth}. From a legal perspective they do not necessarily constitute a novelty, since most European States, including Poland, criminalize the abuse of freedom of expression (which includes political speech) or impose other forms of liability \cite{Rosenfeld2003HateAnalysis}. The Polish Election Code of 2011 \cite{PEC} requires the oversight of the National Elections Committee (\textit{Państwowa Komisja Wyborcza}, hereinafter ``NEC''), a supreme and permanent body charged with overseeing the implementation of electoral law. 
Unfortunately, the Polish election administration lacks the ability to track the spread and quantity of messages that count as electoral agitation (electioneering) under the Polish Election Code. In this paper, we present a data set and model that allows us to identify not just any political speech on Polish Twitter but precisely content categorized as regulated political speech in Polish law. It may provide valuable information, e.g., on the number of posts 
campaigning in favor of a specific candidate, the number of breaches of pre-election silence, as well as the vast number of data used for political science, journalists, international observers, and other parties following national politics. In the long term, it may help courts and the NEC to verify whether election campaigns comply with free and fair election standards (required by the Polish Constitution and article 3 of the 1st Protocol to the European Convention on Human Rights ``right to free and fair elections''). 

\vspace{0.25em}

Some natural language resources addressing political content analysis in social media already exist. These include collections related to elections in countries such as Spain \cite{Taule2018a}, France \cite{Lai2019b}, and Italy \cite{Lai2018}. While the data sets on political campaigning are fundamental for studies on social media manipulation \cite{Aral2019ProtectingManipulation}, there are very limited resources \cite{augustyniak} that allow us to understand the agitation phenomena in Polish. Thus, there is a strong need to acquire data annotated in accordance with legal conditions. We want to fill this gap and present a textual data set of agitation during the 2020 Polish Presidential Election that was annotated in terms of precise, legally conditioned categories. Our contributions are as follows: (1) a novel, publicly open data set for detecting electoral agitation in the Polish language, (2) a publicly available neural-based model for classifying social media content with electioneering that achieves a 68\% F1 score, and (3) an 
analysis of the agitation during the Polish 2020 Presidential Election campaign.

\section{Motivation and Legal Framework}
In the 21 July 2009 (case no. K 7/09) ruling, the Polish Constitutional Tribunal explained that free elections constitute the core element of the rule of law creating a positive obligation for parliament ``to establish rules which provide citizens with access to truthful information on public matters and about candidates. The election campaign shall lead to a free formation of electorate will and conclusive decision expressed by an act of voting''. What rivets the attention of the doctrine is the Tribunal's emphasis on a certain quality of information (``truthful'') and its unconstrained exchange ``by all citizens''. This demand for an inclusive, democratic debate creates a difficult situation when it comes to balancing the freedom of expression and the demand for truthful, regulated political campaigns on social media platforms - a medium that was practically unknown at the time of issuing those judgments. 
In Poland, electioneering may commonly be called ``political advertising'', a form of corporate jargon originating from areas such as Twitter’s or Facebook’s terms of service. Primarily, however, it has a normative definition in Article 105 of the Election Code. 
The Election Code defines electioneering as ``public encouraging to vote a certain way or for a candidate of the election committee'' (Article 105 sec. 1 EC). Public posts or tweets fall under this category, but it is still unclear whether ``public'' would apply to political micro- or nanotargeting as long as they remain personalized messages targeting individual recipients (the existing case law suggests that the adjective ``public'' in electoral agitation may not refer exclusively to ``open, unrestricted'' access to political ads but also to a message distributed to a ``considerable'' number of recipients, regardless of possible personalization by means of microtargeting). 
During the 2019 parliamentary campaign, the OSCE Office for Democratic Institutions and Human Rights pointed out the lack of an official monitoring mechanism \cite{OSCE}. Addressing this research gap, we believe our data set helps to identify tweets that contain public encouraging, as well as those trying to target election participation in general. 

\section{Data Set Creation Process}
\subsection{Data collection}

The data set was obtained from the social network Twitter and contains 9,819,490 tweets. They come from the 2020 presidential election campaign, which was defined by the official time frame from 05.02.2020 to 12.07.2020. The texts were collected based on the following hashtags: \textit{biedron2020}, \textit{bosak2020}, \textit{druzynakosiniaka}, \textit{czasdecyzji}, \textit{duda2020}, \textit{ekipaszymona}, \textit{głosuj}, \textit{hołownia2020}, \textit{idziemynawybory}, \textit{kosiniak2020}, \textit{kosiniakkamysz}, \textit{mimowszystkoduda}, \textit{NieKłamRafał}, \textit{pis}, \textit{po}, \textit{polska2050}, \textit{PrezydentRP}, \textit{rafał}, \textit{rafałniekłam}, \textit{trzaskowski2020}, \textit{wybory}, \textit{wybory2020}, \textit{wyboryprezydenckie2020}, \textit{wyboryprezydenckie}, \textit{wypad}. Duplicated texts (retweets) and tweets in languages other than Polish were filtered out, which resulted in 4,952,804 tweets. We also excluded all tweets shorter than 100 characters (without counting mentions, links and hashtags, to make each text likely to be more informative). After consulting with domain experts, tweets containing a hashtag related to a candidate (e.g. \textit{\#Duda2020}, \textit{\#Trzaskowski2020}) were also filtered out, as such hashtags that directly point to candidates and most likely indicate a direct link to electoral agitation activity. Finally, after careful data cleaning, the number of samples in the data set was 15,790 tweets, which included hashtags with a neutral tinge (without a direct candidate or party mention). This set was prepared as the main corpus for annotation.

\subsection{Annotation method}
In order to solve the problem of detecting electoral agitation, texts can be divided into two categories: agitated and not agitated. However, relying directly on the definition of electoral agitation in Article 105 of the Polish Election Code and analysis of data characteristics, we adopted four mutually exclusive categories for the annotation:

\vspace{0.25em}

\textbf{Inducement} – explicitly convincing or advising voting for/against a particular candidate. It must be clear from the content which person the text is about and it may also include tweets regarding a political party. This is the most significant category, which focuses on agitation in the intuitive sense – e.g.: ``Vote for Duda!''.

\textbf{Encouragement} – assigning good or bad associations and characteristics to a person. This category includes tweets that refer to candidates or parties directly but are not agitation in a literal sense. These are usually positive or negative statements about politicians – e.g.: ``Holownia makes a good impression, maybe he will be a new hope for Poland.''.

\textbf{Voting turnout} – refers to encouraging or discouraging people from participating in the vote itself, e.g.: ``Poles, go vote!''. 

\textbf{Normal} – non-agitated text or text that does not qualify as the other groups. A neutral statement or discussion piece that does not fall into any of the above groups. e.g.: ``I'm curious about this election, the fight will go on until the very end.''.

\vspace{0.25em}

The \textit{inducement} and the \textit{encouragement} categories are both considered as electoral agitation. In contrast, encouraging people to participate in voting itself (without intending to influence the voter's decision) is not treated as agitation based on current legal regulations in the Electoral Code. The correctness of the selection and interpretation of the above categories was confirmed by a constitutional and legal field expert. The categories chosen in the presented way cover the most relevant needs for interpreting texts in the problem of classifying electoral agitation. However, to extend the data set application, we made an effort to assign tweets with additional metadata that is independent of agitation categories. Each tweet can have multiple additional pieces of information assigned as follows:

\vspace{0.25em}

\textbf{Satire} – a statement that is deemed as satire. Intentionally changing the name of a candidate/party to a satirical term (e.g. Duda – pen, etc.). Various types of rhymes, statements overtly indicating so-called \textit{bait}, which are usually specific to a particular language.

\textbf{Missing context} – the text requires additional knowledge to assign it to a given category. These are often hard-to-catch allusions or statements that refer to a URL or attachment that we don't know about - e.g. ``Let him finish already, I can't stand this steak of nonsense...'' with a URL attached.

\textbf{Media report} – a statement from a media entity, e.g. TV, radio, newspaper. The statement has an informative purpose. Most often it is a quotation, paraphrase or citation of the full statement – e.g. ``\#czasdecyzji – Minister Łukasz Szumowski will be our guest today.''.

\vspace{0.25em}

Before annotation, the data was pre-processed by removing emoticons and hyperlinks as they did not add any relevant information to the target classes. All hashtags and mentions were kept in the text, making them an integral part of the sentences due to the fact that they carry important contextual information.

\vspace{0.25em}

Five native speakers took part in the annotation process. Two annotators labeled each example. We disambiguated and improved the annotation via an additional pass by a third annotator, who resolved the mismatches between the first two annotators and made the data set more consistent and comprehensive. Due to the pioneering nature of the data set being created and concern for its potential usefulness, approaches to interpreting difficult texts during annotation were consulted and coordinated under the supervision of the domain expert. After 4,529 examples had been annotated, when Cohen's kappa score stabilized at a satisfactory level, each sample was labeled by one annotator only. In total, we annotated 6,112 tweets that were randomly selected from the main corpus. 

\begin{table}
\centering
\begin{tabular}{lrrrr}
\toprule
\textbf{Label}          &   \textbf{S} &   \textbf{C}    &   \textbf{M}    &   \textbf{Total} \\
\midrule
Inducement              &   114         &   130     &   17      &   \textbf{735}      \\
Encouragement           &   181         &   356     &   96      &   \textbf{1 517}        \\
Voting turnout          &   5           &   50      &   9       &   \textbf{314}      \\
Normal                  &   156         &   571     &   743     &   \textbf{3 546}    \\
\midrule
\textbf{Total}          &   456        &   1 107     &   865     &   \textbf{6 112}      \\
\bottomrule
\end{tabular}
\caption{The number of examples in the data set. Tweets' metadata: S – satirical, C – missing context, and M – media report.}
\label{tab:labelcounts}
\end{table}

\vspace{0.25em}

Ultimately, we achieved a 0.66 Cohen’s kappa score for four-class annotation. According to \cite{mchugh2012interrater}, it lies between a moderate and strong level of annotation agreement. Table~\ref{tab:labelcounts} shows the counts of annotated tweets per label as well as information on the number of tweets marked as satire, using missing context, and media reports. We conclude that annotating social media content requires additional knowledge or the ability to follow satirical or slang language. This may hinder the learning process of the model and could negatively affect the outcomes.

\subsection{Experiments}

We trained a classifier based on a Polish Language Model called HerBERT \cite{rybak2020klej} that is dedicated to the Polish language. Finally, we achieved a 68\% F1 score for the stratified train-test split in an 80/20 ratio. Table~\ref{tab:labels} presents the precision, recall and F1 score for each label. The data set and trained model are publicly available in the GitHub repository\footnote{
https://github.com/mateuszbaransanok/e-agitation
}.
The outcome of our work is also available at 
\url{www.smart-wust.ml} in the section \textit{Agitation},
where you can reach additional analysis and insights as well as examples of model usage including the scoring of your own entered text. 

\begin{table}
\centering
\begin{tabular}{lccc}
\toprule
\textbf{Label}          &   \textbf{P}          &   \textbf{R}          &   \textbf{F1}\\
\midrule
Inducement              &   71\%                &   52\%                &   60\% \\
Encouragement           &   66\%                &   55\%                &   60\% \\
Voting turnout          &   70\%                &   70\%                &   70\% \\
Normal                  &   74\%                &   86\%                &   80\% \\
\midrule
\textbf{Macro-average}  &   \textbf{70\%}       &   \textbf{66\%}       &   \textbf{68\%} \\
\bottomrule
\end{tabular}
\caption{Precision, Recall, and F1 score for each label of HerBERT model on the agitation data set.}
\label{tab:labels}
\end{table}

\section{Polish Presidential Election - Use Case}

The proposed data set makes it possible to discover electoral agitation in social media. The data set accompanied with the sample model we proposed can help electoral administration and non-governmental bodies to quantitatively analyze the magnitude of the agitation phenomenon. Thus, we performed a study over agitation during the 2020 Presidential Election (tweets classification) to take the share of agitation in the overall political discourse into account. We were interested in the last few weeks of the campaign, where the percentage of agitation tweets was at its highest and pre-election silence was in place, where agitation is prohibited by law.

\begin{figure}[ht!]
    \centering
    \includegraphics[width=0.85\columnwidth]{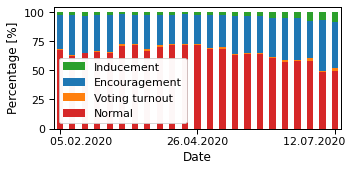}
    \caption{Percentage participation of tweets annotation categories in particular weeks during campaign.}
    \label{fig:campaign}
\end{figure}

\vspace{0.25em}

We performed an experiment on a sample of 1,531,624 untagged tweets, where each tweet was classified using our trained model. The results shown in Figure~\ref{fig:campaign} denote the final stage of the campaign being dominated by agitation (\textit{inducement} and \textit{encouragement} increased in total of 17 pp). More than a third of tweets contained agitation in the presidential election discourse. The encouragement to participate in voting increased in the weeks leading up to the day of the election, but nevertheless, they constituted a clear minority.

\begin{figure}[ht!]
    \centering
    \includegraphics[width=0.85\columnwidth]{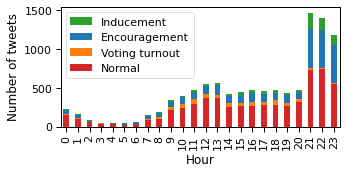}
    \caption{Number of tweets per hour during second round election day (12.07.2020). The pre-election silence ended at 9 p.m (21:00).}
    \label{fig:silence}
\end{figure}

\vspace{0.25em}

We also spotted the problem of electoral agitation on Twitter during pre-election silence, as shown in Figure~\ref{fig:silence}.
This also confirms \cite{MusiaKarg2018TheES} the doubts whether the ban on campaigning during pre-election silence even exists, since so many Twitter users campaign for politicians during that time, violating the existing law.
Even though the agitation decreases threefold in comparison to periods outside of pre-election silence, it still exists and may affect voters as a result. 
To the best of our knowledge, the result is a first attempt to automatically discover agitation in social media in Poland. Above all, further research and method development provide the prospect of supporting the judiciary in ensuring the fairness of the election campaign and freeing social media from political propaganda.

\begin{figure}[ht!]
    \centering
    \includegraphics[width=0.46\textwidth]{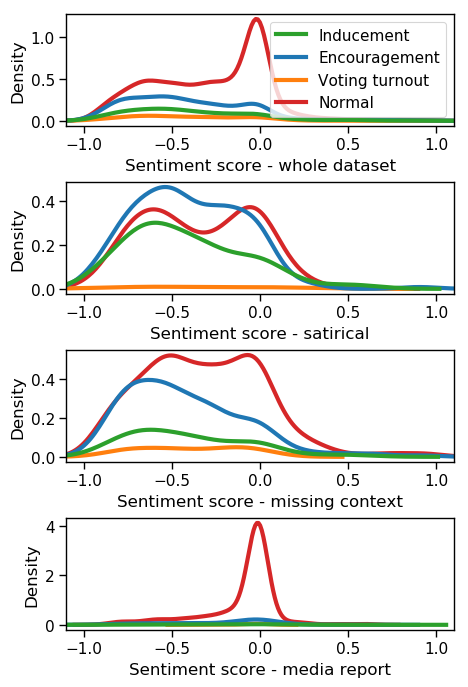}
    \caption{Sentiment score probability density function for all annotation classes with metadata classes distinction.}
    \label{fig:sentiment}
\end{figure}

\vspace{0.25em}

The data set we propose also enabled us to analyze sentiment polarity across categories. Figure~\ref{fig:sentiment} shows the estimated sentiment probability density function with respect to all metadata labels. The sentiment was assigned to a $[-1,1]$ (negative to positive) range based on \cite{strzalka2020sentimentpl}. 
As we can see, even though \textit{normal} content has mixed sentiment, it is mostly neutral. Posts of an agitating nature are completely different where sentiment is very negative, especially in the \textit{encouragement} category. This proves that social media campaigns are mainly based on negative narration \cite{Piontek2017} and frequently take the form of hate-speech. The least polarized class is the \textit{voting turnout} where sentiment is uniformly balanced relative to all classes. As anticipated, media news falls mainly into the \textit{normal} category and is characterized by a strictly neutral sentiment, which is the case even when media texts are classified as electoral agitation. Our analyses indicate that under election campaign conditions, the sentiment of a text itself can carry valuable information about the intentions behind it, something especially evident for unbiased media statements confronted with satirical content.

\section{Conclusions and Future Work}
The presented data set along with the model enables us to categorize social media posts in terms of electoral agitation and evaluate the number of posts campaigning in favor of candidates or the number of times pre-election silence was breached. This makes it suitable for the needs of a wide range of audiences, from researchers to journalists, electoral committees, and government authorities that care about the integrity of elections. The usefulness of studying this phenomenon is underlined by the presented case studies. Based on the collected data we can describe the studied election campaign as gaining in agitation intensity over time and highly offensive in the case of social media content. Publishing this data set makes it possible to train modern NLP models with various applications in the area of policy and law. We expect that the publication of this data set and its future use, in conjunction with machine learning techniques, will lead to increased fairness in election processes and will help reduce the spread of propaganda.

\vspace{0.25em}

In this paper, we have identified promising areas of research using the composed data set. We plan to work on more data sets and models to ensure the integrity of the political campaigns, classify electoral agitation content, and widen natural language solutions in regards to the content in social media.

\section{Bibliographical References}\label{reference}

\bibliographystyle{lrec2022-bib}
\bibliography{lrec2022}


\end{document}